\documentclass{article}
\usepackage{spconf,amsmath,graphicx}
\usepackage{times}
\usepackage{url}
\usepackage{latexsym}
\usepackage{multirow}
\usepackage{amssymb, bm}
\usepackage{arydshln}

\newcommand{\newcite}[2]{#1 et al.~\cite{#2}}
\newcommand{\parjump}{\vspace{+0.5em}}
\newcommand{\newparagraph}[1]{\par \parjump \noindent \textbf{#1}}

\title{A Co-Interactive Transformer for Joint Slot Filling and Intent Detection}
\name{Libo Qin$^{\star}$ \qquad Tailu Liu$^{\star}$ \qquad Wanxiang Che$^{\dagger}$ \qquad Bingbing Kang \qquad Sendong Zhao \qquad Ting Liu
\thanks{$^{\star}$ Equal contributions.}
\thanks{$^{\dagger}$ Corresponding author.}}
\address{Research Center for Social Computing and Information Retrieval, Harbin Institute of Technology, China}
\begin{document}
%
\maketitle

\begin{abstract}
Intent detection and slot filling are two main tasks for building a spoken language understanding (SLU) system. 
The two tasks are closely related and the information of one task can benefit the other.
Previous studies either implicitly model the two tasks with multi-task framework or only explicitly consider the single information flow from intent to slot.
None of the prior approaches model the bidirectional connection between the two tasks simultaneously in a unified framework.
In this paper, we propose a Co-Interactive Transformer which considers the cross-impact between the two tasks. 
Instead of adopting the self-attention mechanism in vanilla Transformer, we propose a co-interactive module
to consider the cross-impact by building a bidirectional connection between the two related tasks, where slot and intent can be able to attend on the corresponding mutual information.
The experimental results on two public datasets show that our model achieves the state-of-the-art
performance.
\end{abstract}

\begin{figure*}[t]
	\centering
	\includegraphics[width=0.75\textwidth]{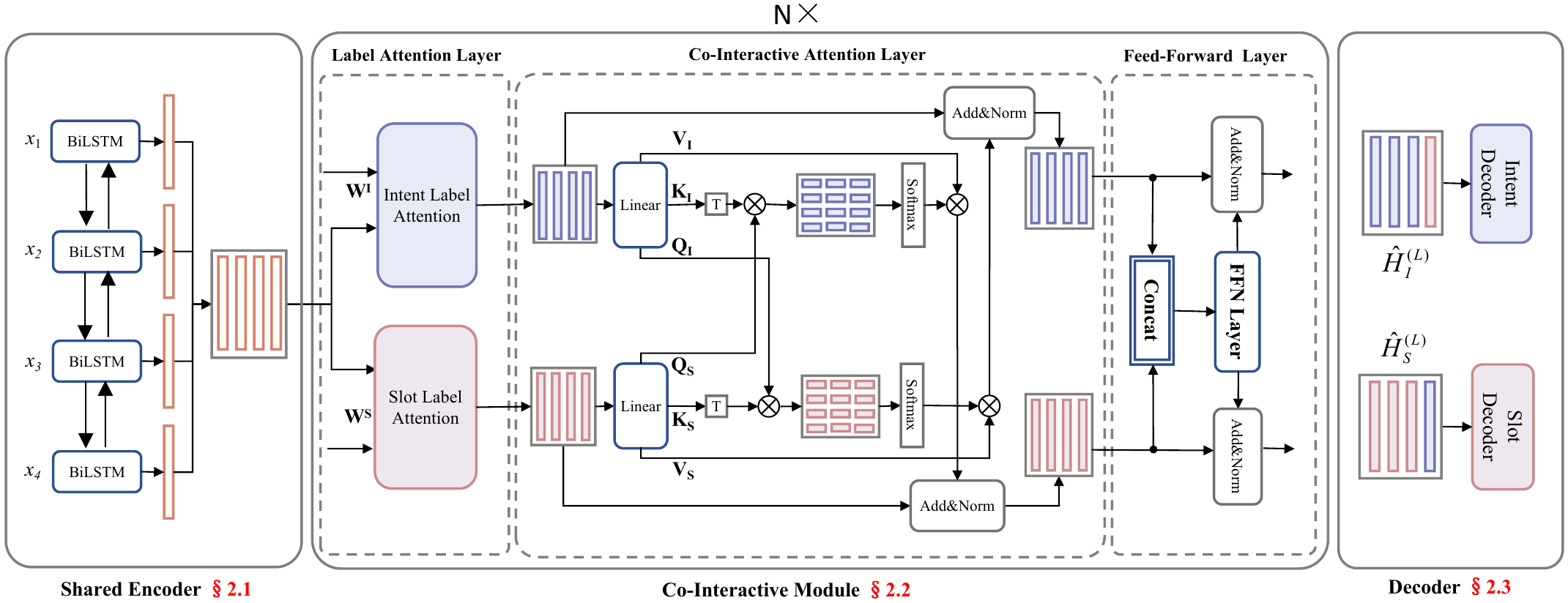}
	\caption{The illustration of the co-interactive transformer.}
	\label{fig:framework}
\end{figure*}
\section{Introduction}
\label{sec:intro}
\textit{Spoken language understanding (SLU)} typically consists of two typical subtasks including intent detection and slot filling, which is a critical component in task-oriented dialogue systems \cite{tur2011spoken}.
For example, given ``\textit{watch action movie}'', intent detection can be seen an classification task to identity an overall intent class label (i.e., \texttt{WatchMovie}) and slot filling can be treated as a sequence labeling task to produce a slot label sequence (i.e., \texttt{O}, \texttt{B-movie-type}, \texttt{I-movie-type}).
Since slots and intent are highly closed, dominant SLU systems in the literature \cite{liu2016attention,zhang2016joint,goo2018slot,li2018self,qin-etal-2019-stack} proposed joint model to consider the correlation between the two tasks. 
Existing joint models can be classified into two main categories. 
The first strand of work \cite{liu2016attention,zhang2016joint} adopted a multi-task framework with a shared encoder to solve the two tasks jointly.
While these models outperform the pipeline models via mutual enhancement, they just modeled the relationship implicitly by sharing parameters.  
The second strand of work \cite{goo2018slot,li2018self,qin-etal-2019-stack} explicitly applied the intent information to guide the slot filling task and achieve the state-of-the-art performance. 
However, they only considered the single information flow from intent to slot.

We consider addressing the limitation of existing
works by proposing a Co-Interactive Transformer for joint slot filling and intent detection.
Different from the vanilla Transformer \cite{NIPS2017_7181}, 
the core component in our framework is a proposed co-interactive module to model the relation between the two tasks, aiming to consider the cross-impact of the two tasks and  enhance the two tasks in a mutual
way. 
Specifically, in each co-interactive module, we first apply the label attention mechanism \cite{cui-zhang-2019-hierarchically} over intent and slot label to capture the initial \textit{explicit intent and slot representations}, which extracts the intent and slot semantic information.
Second, the explicit intent and slot representations are fed into a co-interactive attention layer to make mutual interaction.
In particular, the \textit{explicit intent representations} are treated as queries and slot representations are considered as keys as well as values to obtain the slot-aware intent representations.
Meanwhile, the \textit{explicit slot representations} are used as queries and intent representations are treated as keys as well as values to get the intent-aware slot representations.
These above operations can establish the bidirectional connection across intent and slots. The underlying intuition is that slot and intent can be able to attend on the corresponding mutual information with the co-interactive attention mechanism.

The experimental results on two benchmarks SNIPS \cite{coucke2018snips} and ATIS \cite{goo2018slot} show that our framework achieves significant improvement compared to all baselines.
In addition, we incorporate the pre-trained model (BERT)
\cite{devlin2018bert} in our framework, which can achieve a new state-of-the-art performance.
Code for this paper are publicly available at
\url{https://github.com/kangbrilliant/DCA-Net}.

\section{Approach}
\label{sec:approach}
This section describes the details of our framework.
As shown in Figure~\ref{fig:framework}, it mainly consists of
a shared encoder ($\S\ref{sec:shared_encoder}$), a co-interactive module ($\S\ref{sec:co-interactive}$) that explicitly establishes bidirectional connection between the two tasks, and two separate decoders ($\S\ref{sec:decoders}$) for intent detection and slot filling.

\subsection{Shared Encoder}\label{sec:shared_encoder}
We use BiLSTM \cite{hochreiter1997long} as the shared encoder, which aims to leverage the advantages of temporal features within word orders.
BiLSTM consists of two LSTM layers.
For the input sequence $\{{{x}}_{1}, {{x}}_{2}, \ldots, {{x}}_{n}\}$ ($n$ is the number of tokens.), BiLSTM reads it
forwardly and backwardly to produce a series of context-sensitive hidden states $\mathbf{H} = \{\mathbf{h}_{1}, \mathbf{h}_2, \ldots, \mathbf{h}_{n}\}$ by repeatedly applying the recurrence $\mathbf{h}_{i}$ = BiLSTM ($\phi^{emb}(x_{i})$, $\mathbf{h}_{i-1}$), where $\phi^{emb}(\cdot)$ represents the embedding function.
\subsection{Co-Interactive Module}\label{sec:co-interactive}
The Co-Interactive module is the core component of our framework, aiming to build the bidirectional connection between intent detection and slot filling.

In vanilla Transformer, each sublayer consists of a self-attention and a feed-forward network  (FFN) layer.
In contrast, in our co-interactive module, we first apply a intent and slot label attention layer to obtain the explicit intent and slot representation.
Then, we adopt a co-interactive attention layer instead of self-attention to model the mutual interaction explicitly.
Finally, we extend the basic FFN for further fusing intent and slot information in an implicit method.
\subsubsection{Intent and Slot Label Attention Layer}
Inspired by \newcite{Cui}{cui-zhang-2019-hierarchically} that successfully captures label representations, we
perform label attention over intent and slot label to get the explicit intent representation and slot representation. Then, they are fed into co-interactive attention layer to make a mutual interaction directly.
In particular, we use the parameters of the fully-connected slot filling decoder layer and intent detection decoder layer as slot embedding matrix $\mathbf{{W}^{S}}\in \mathbb{R}^{d\times \left | \mathbf{S}^{label} \right |}$ and intent embedding matrix $\mathbf{{W}^{I}}\in \mathbb{R}^{d\times \left | \mathbf{I}^{label} \right |}$ ($d$ represents the hidden dimension; $\left | \mathbf{S}^{label} \right |$ and $\left | \mathbf{I}^{label} \right |$ represents the number of slot and  intent label, respectively), which can be regarded as the distribution of labels in a certain sense.
\newparagraph{Intent and Slot Representations}
In practice,  we use $\mathbf{H}\in \mathbb{R}^{n\times d }$ as the query, $\mathbf{{W}^{v}}\in \mathbb{R}^{d\times \left | \mathbf{v}^{label} \right |}$  ($\mathbf{v}$ $\in$ \{$\mathbf{I}$ or $\mathbf{S}$\})  as the key and value to obtain intent representations $\mathbf{{H}_{v}}$ with intent label attention:
\begin{align}
\mathbf{A}&=\operatorname{softmax}( \mathbf{H} \mathbf{{W}^{v}}), \\
\mathbf{H_v}&=\mathbf{H}+\mathbf{A}\mathbf{{W}^{v}},
\end{align}
where $\mathbf{I}$ denotes the intent and $\mathbf{S}$ represents the slot.

Finaly, $\mathbf{{H}_{I}}\in \mathbb{R}^{n\times d }$ and $\mathbf{{H}_{S}}\in \mathbb{R}^{n\times d }$ are the obtained explicit intent representation and slot representation, which capture the intent and slot semantic information, respectively.
\subsubsection{Co-Interactive Attention Layer}
$\mathbf{{H}_{S}} $ and $\mathbf{{H}_{I}}$ are further used in next co-interactive attention layer to model mutual interaction between the two tasks.
This makes the slot representation updated with the guidance of associated intent and intent representations updated with the guidance of associated slot, achieving a bidirectional connection with the two tasks.

\newparagraph{Intent-Aware Slot and Slot-Aware Intent Representation}
Same with the vanilla Transformer, we map the matrix $\mathbf{{H}_{S}}$ and $\mathbf{{H}_{I}}$ to queries ($\mathbf{{Q}_{S}}$, $\mathbf{{Q}_{I}}$), keys ($\mathbf{{K}_{S}}$, $\mathbf{{K}_{I}}$) and values ($\mathbf{{V}_{S}}$, $\mathbf{{V}_{I}}$) matrices by using different linear projections.
To obtain the slot representations to incorporate the corresponding intent information, it is necessary to align slot with its closely related intent.  We treat $\mathbf{{Q}_{S}}$ as queries, $\mathbf{{K}_{I}}$ as keys and $\mathbf{{V}_{I}}$ as values.
The output is a weighted sum of values:
\begin{align}
{\mathbf{C}}_\mathbf{S} &= \operatorname { softmax } \left( \frac { \mathbf {Q}_\mathbf{S} \mathbf K_\mathbf{I}^\top } { \sqrt { d _ { k } } } \right) \mathbf {{V}_{I}}, \\
\mathbf{H}^{'}_\mathbf{S}&= \mathbf{LN}\left(\mathbf{H_S}+ \mathbf{C_S}\right),\end{align}
where $\mathbf{LN}$ represents the layer normalization function \cite{ba2016layer}.

Similarly, we treat $\mathbf{{Q}_{I}}$ as queries, $\mathbf{{K}_{S}}$ as keys and $\mathbf{{V}_{S}}$ as values to obtain the slot-aware intent representation $\mathbf{H}^{'}_\mathbf{I}$.
$\mathbf{H}^{'}_\mathbf{S}\in \mathbb{R}^{n\times d }$ and  $\mathbf{H}^{'}_\mathbf{I}\in \mathbb{R}^{n\times d }$ can be considered as leveraging the corresponding slot and intent information, respectively.
\subsubsection{Feed-forward Network Layer}
In this section, we extend feed-forward network layer to implicitly fuse intent and slot information.
We first concatenate
$\mathbf{H_I}'$ and $\mathbf{H_S}'$  to combine the slot and intent information.
\begin{align}
\mathbf{{H}}_\mathbf{IS}&=\mathbf{H}_\mathbf{I}' \oplus \mathbf{H}_\mathbf{S}',
\end{align}
where $	\mathbf{H_{IS}}$ = ($	\mathbf{{h}}^{1}_\mathbf{IS}$, $	\mathbf{{h}}^{2}_\mathbf{IS}$,$\dots$, $\mathbf{{h}}^{n}_\mathbf{IS}$) and $\oplus$ is concatenation.

Then, we follow \newcite{Zhang}{zhang2016joint} to use word features for each token, which is formated as:
\begin{align}
\mathbf{h}_{(f,t)}^{t}&=\mathbf{h}_\mathbf{IS}^{t-1}\oplus \mathbf{h}_\mathbf{IS}^{t} \oplus \mathbf{h}_\mathbf{IS}^{t+ 1}.
\end{align}

Finally, FFN layer fuses the intent and slot information:
\begin{align}
\mathbf{FFN}(\mathbf{H}_{(f,t)})&=\operatorname{max}(0,	\mathbf{H}_{(f,t)}\mathbf{W}_1+b_1)\mathbf{W}_2+b_2,\\
\hat{\mathbf{H_I}} &=\mathbf{LN}(\mathbf{H}_\mathbf{I}^{'} +\mathbf{FFN}(\mathbf{H}_{(f,t)})),\\
\hat{\mathbf{H_S}} &=\mathbf{LN}(\mathbf{H}_\mathbf{S}^{'} +\mathbf{FFN}(\mathbf{H}_{(f,t)})),
\end{align}
where $\mathbf{H}_{(f,t)}=(\mathbf{h}_{(f,t)}^{1},\mathbf{h}_{(f,t)}^{2},...,\mathbf{h}_{(f,t)}^{t})$; $ \hat{\mathbf{H_I}}$ and $ \hat{\mathbf{H_S}}$ is the obtained updated intent and slot information that aligns corresponding slot and intent features, respectively.

\subsection{Decoder for Slot Filling and Intent Detection}\label{sec:decoders}
In order to conduct sufficient interaction between the two tasks, we apply a stacked co-interactive attention network with multiple layers.
After stacking $L$ layer, we obtain a final updated slot and intent representations
$\hat{\mathbf{H}}^\mathbf{(L)}_\mathbf{I} = (\hat{\mathbf{h}}^\mathbf{(L)}_{(\mathbf{I},1)}, \hat{\mathbf{h}}^\mathbf{( L)}_{(\mathbf{I},2)}, ..., \hat{\mathbf{h}}^\mathbf{(L)}_{(\mathbf{I},n)}), \hat{\mathbf{H}}^\mathbf{( L)}_\mathbf{S} (\hat{\mathbf{h}}^\mathbf{(L)}_{(\mathbf{S},1)}, \hat{\mathbf{h}}^\mathbf{( L)}_{(\mathbf{S},2)}, ..., \hat{\mathbf{h}}^\mathbf{( L)}_{(\mathbf{S},n)})$.
\newparagraph{Intent Detection}
We apply \textit{maxpooling} operation \cite{kim-2014-convolutional} on $\hat{\mathbf{H}}^\mathbf{(L)}_\mathbf{I} $ to obtain sentence representation $\mathbf{c}$, which
is used as input for intent detection:
\begin{eqnarray}
{\hat{\mathbf{y}}} ^ \mathbf{ I } &=& \operatorname { softmax } \left( {\mathbf{W}} ^ \mathbf{ I }{{\mathbf{c}}+{\mathbf{b}}_\mathbf{S}}  \right), \\
{o}  ^ \mathbf{ I } &=& \operatorname { argmax } ({\mathbf{y}}  ^ \mathbf{ I }),
\end{eqnarray}
where ${\hat{\mathbf{y}}} ^ \mathbf{ I }$ is the output intent distribution; ${{o}^\mathbf{I}} $ represents the intent label and ${\mathbf{W}}^\mathbf{I}$ are trainable parameters of the model.

\newparagraph{Slot Filling}
We follow \newcite{E}{e-etal-2019-novel} to apply a standard CRF layer
to model the dependency between labels, using:
\begin{eqnarray}\label{eq:crf}
\mathbf{O_S} &=& \mathbf {W^S}\mathbf{\hat{H}}^\mathbf{(L)}_\mathbf{S} + \mathbf{b}_\mathbf{S},\\
P(\hat{\mathbf{y}}| \mathbf{O_S})&=&\frac{\sum_{i=1}\exp f(y_{i-1},y_i,\mathbf{O_S})}{\sum_{y^\prime}\sum_{i=1}\exp f(y^\prime_{i-1},y^\prime_i,\mathbf{O_S})},
\end{eqnarray}
where $f(y_{i-1},y_i,\mathbf{O}_\mathbf{S})$ computes the transition score from $y_{i-1}$ to $y_{i}$ and $\hat{\mathbf{y}}$ represents the predicted label sequence.

\begin{table*}[t]
	\centering
	\resizebox{0.7\textwidth}{!}{
		\begin{tabular}{l|ccc|ccc}
			\hline
			\multirow{2}{*}{\textbf{Model}} & \multicolumn{3}{c|}{\textbf{SNIPS}} &\multicolumn{3}{c}{\textbf{ATIS}}\\ \cline{2-7}
			&\textbf{Slot (F1)}&\textbf{Intent (Acc)}&\textbf{Overall (Acc)}&\textbf{Slot (F1)}&\textbf{Intent (Acc)}&\textbf{Overall (Acc)}\\
			\hline
			Slot-Gated Atten \cite{goo2018slot}  & 88.8& 97.0& 75.5& 94.8& 93.6& 82.2\\
			SF-ID Network \cite{e-etal-2019-novel}   &90.5& 97.0& 78.4& 95.6& 96.6& 86.0\\
			CM-Net \cite{liu-etal-2019-cm}   &93.4& 98.0& 84.1& 95.6& 96.1& 85.3\\
			Stack-Propagation \cite{qin-etal-2019-stack}& 94.2& 98.0& 86.9& 95.9& 96.9& 86.5\\
			\hdashline
			Our framework &\textbf{95.9}&\textbf{ 98.8}& \textbf{90.3}& \textbf{95.9}& \textbf{97.7}& \textbf{87.4}\\
			\hline
			Stack-Propagation + BERT \cite{qin-etal-2019-stack}&	97.0 &{99.0}& 92.9& 96.1& 97.5& 88.6	\\
			
			Our framework + BERT &{97.1}&{98.8}&{93.1}&	{96.1}&	{98.0}&	{88.8}\\
			\hline
	\end{tabular}}
	\caption{\label{sec:main_results} Slot filling and intent detection results on two datasets. 
	}
\end{table*}
\begin{table*}[t]
	\centering
	\small
	\resizebox{0.7\textwidth}{!}{
		\begin{tabular}{l|ccc|ccc}	
			\hline
			\multirow{2}{*}{\textbf{Model}} & \multicolumn{3}{c|}{\textbf{SNIPS}} &\multicolumn{3}{c}{\textbf{ATIS}}\\ \cline{2-7}
			&\textbf{Slot (F1)}&\textbf{Intent (Acc)}&\textbf{Overall (Acc)}&\textbf{Slot (F1)}&\textbf{Intent (Acc)}&\textbf{Overall (Acc)}\\
			\hline
			without intent attention layer	&95.8&	98.5&	90.1&	95.6&	97.4&	86.6\\
			without slot attention layer&	95.8	&98.3&	89.4&	95.5	&97.6&	86.7\\
			self-attention mechanism&	95.1&	98.3&	88.4	&95.4&	96.6	&86.1\\
			with intent-to-slot	&95.6&	98.4&	89.3&	95.8&	97.1	&87.2\\
			with slot-to-intent	&95.4&	98.7&	89.4&	95.5&	97.7	&87.0\\
			\hline
			Our framework	&\textbf{95.9}&\textbf{98.8}&	\textbf{90.3}&	\textbf{95.9}&	\textbf{97.7}&	\textbf{87.4}\\
			
			\hline
		\end{tabular}
	}
	\caption{\label{sec:ablation} Ablation experiments on the SNIPS and ATIS datasets. }
\end{table*}
\section{Experiments}
\label{sec:experiments}

\subsection{Dataset}
We conduct experiments on two benchmark datasets.
One is the public ATIS dataset \cite{hemphill1990atis} and another is SNIPS \cite{coucke2018snips}.
Both datasets are used in our paper following the same format and partition as in \newcite{Goo}{goo2018slot} and \newcite{Qin}{qin-etal-2019-stack}.

In the paper, the hidden units of the shared encoder and the co-interactive module are set as 128. We use 300d GloVe pre-trained vector \cite{Pennington2014Glove} as the initialization embedding. The number of co-interactive module is 2. L2 regularization used on our model is $1\times 10^{-6}$ and the dropout ratio of co-interactive module is set to 0.1.
We use Adam \cite{kingma-ba:2014:ICLR} to
optimize the parameters in our model.

Following \newcite{Goo}{goo2018slot} and \newcite{Qin}{qin-etal-2019-stack},
intent detection and slot filling are optimized simultaneously via a joint learning scheme.
In addition, we evaluate the performance of slot filling using F1 score, intent prediction using accuracy, the sentence-level semantic frame parsing using overall accuracy.

\subsection{Main Results}\label{sec:main_result}
Table~\ref{sec:main_results} shows the experiment results.
We have the following observations:
1) Compared with baselines \textit{Slot-Gated} and \textit{Stack-Propagation} that only leverage intent information to guide the slot filling, our framework gain a large improvement.
The reason is that our framework consider the cross-impact between the two tasks where the slot information can be used for improving intent detection.
It's worth noticing that the parameters between our model and \textit{Stack-Propagation} is of the same magnitude, which further verifies that contribution of our model comes from the bi-directional interaction rather than parameters factor.
2) \textit{SF-ID Network} and \textit{CM-Net} also can be seen as considering the mutual interaction between the two tasks.
Nevertheless, their models cannot model the cross-impact simultaneously, which limits their performance. Our framework outperforms \textit{CM-Net} by 6.2\% and 2.1\% on overall acc on SNIPS and ATIS dataset, respectively.
We think the reason is that
our framework achieves the bidirectional connection simultaneously in a unified network.
3) \textit{Our framework + BERT} outperforms the  \textit{Stack-Propagation +BERT}, which verifies the effectiveness of our proposed model whether it's based on BERT or not.

\subsection{Analysis}
\newparagraph{Impact of Explicit Representations} We remove the intent attention layer and replace $\mathbf{{H}_{I}}$ with $\mathbf{H}$. This means that we only get the slot representation explicitly, without the intent semantic information. We name it as \textit{without intent attention layer}. Similarly,
we perform the \textit{without slot attention layer} experiment.
The result is shown in Table~\ref{sec:ablation},
 we observe that the slot filling and intent detection performance drops, which demonstrates the initial explicit intent and slot representations are critical to the co-interactive layer between the two tasks.
\newparagraph{Co-Interactive Attention vs. Self-Attention Mechanism} We use the self-attention layer in the vanilla Transformer instead of the co-interactive layer in our framework, which can be seen as no explicit interaction between the two tasks. Specifically, we concatenate the $\mathbf{{H}_{S}}$ and $\mathbf{{H}_{I}}$ output from the label attention layer as input, which is fed into the self-attention module.
The results are shown in Table \ref{sec:ablation}, we observe that our framework outperforms the \textit{self-attention mechanism}. The reason is that \textit{self-attention mechanism} only model the interaction implicitly while our co-interactive layer can explicitly consider the cross-impact between two tasks.
\newparagraph{Bidirectional Connection vs. One Direction Connection}
We only keep one direction of information flow from intent to slot or slot to intent.
We achieve this by only using one type of information representation
 representation as queries to attend another information representations. We name it as \textit{with intent-to-slot} and  \textit{with slot-to-intent}.
 From the results in Table~\ref{sec:ablation}.
We observe that our framework outperforms \textit{with intent-to-slot} and \textit{with slot-to-intent}.
We attribute it to the reason that modeling the mutual interaction between slot filling and intent detection can enhance the two tasks in a mutual
way. In contrast, their models only consider the interaction from single direction of information flow.

\section{Related Work}
\label{sec:related_work}
Different classification methods, such as support vector machine (SVM) and RNN \cite{haffner2003optimizing,sarikaya2011deep},
 have been proposed to solve Intent detection.
Meanwhile, 
the popular methods are conditional random fields (CRF) \cite{raymond2007generative} and recurrent neural networks (RNN) \cite{xu2013convolutional,yao2014spoken} are proposed to solve slot filling task. 

Recently, many dominant joint models \cite{liu2016attention,zhang2016joint,goo2018slot,li2018self,zhao2019neural,qin-etal-2019-stack,zhang-etal-2018-neural}  are proposed to consider the closely correlated relationship between two correlated tasks.
The above studies either adopt a multi-task framework to model the relationship between slots and intent implicitly or leverages intent information to guide slot filling tasks explicitly.
Compared with their models, 
we propose a co-interactive transformer framework, which simultaneously  considers the cross-impact and establish a directional connection between the two tasks while they only consider the single direction information flow or implicitly model the relationship into a set of shared parameters.

Meanwhile, \newcite{Wang}{wang2018bi}, \newcite{E}{e-etal-2019-novel}, and \newcite{Liu}{liu-etal-2019-cm}
propose models to promote slot filling and intent detection via mutual interaction.
Compared with their methods, the main differences are as following:
1) \newcite{E}{e-etal-2019-novel} introduce a SF-ID network, which includes two sub-networks iteratively achieve the flow of information between intent and slot. Compared with their models, our framework build a bidirectional connection between the two tasks  simultaneously in an unified framework while their frameworks must consider the iterative task order.
2) \newcite{Liu}{liu-etal-2019-cm} propose a collaborative memory block to implicitly consider the mutual interaction between the two tasks, which limits their performance. 
In contrast, our model proposes a co-interactive attention module to explicitly establish the bidirectional connection in a unified framework.

\section{Conclusion}
\label{sec:conclusion}

In our paper, we proposed a co-interactive transformer for joint model slot filling and intent detection, which enables to fully take the advantage of the mutual interaction knowledge.
Experiments on two datasets show the effectiveness of the proposed models and our framework achieves the
state-of-the-art performance. 
\section{Acknowledgements}
\label{sec:acknowledgements}
This work was supported by the National Key R\&D Program of China via grant 2020AAA0106501 and the National Natural Science Foundation of China (NSFC) via grant 61976072 and 61772153.
This work was supported by the Zhejiang Lab's International Talent Fund for Young Professionals.

%


\bibliographystyle{IEEEbib}
\bibliography{refs}

\begin{thebibliography}{10}

\bibitem{tur2011spoken}
Gokhan Tur and Renato De~Mori,
\newblock {\em Spoken language understanding: Systems for extracting semantic
  information from speech},
\newblock John Wiley \& Sons, 2011.

\bibitem{liu2016attention}
Bing Liu and Ian Lane,
\newblock ``Attention-based recurrent neural network models for joint intent
  detection and slot filling,''
\newblock {\em arXiv preprint arXiv:1609.01454}, 2016.

\bibitem{zhang2016joint}
Xiaodong Zhang and Houfeng Wang,
\newblock ``A joint model of intent determination and slot filling for spoken
  language understanding.,''
\newblock in {\em Proc. of IJCAI}, 2016.

\bibitem{goo2018slot}
Chih-Wen Goo, Guang Gao, Yun-Kai Hsu, Chih-Li Huo, Tsung-Chieh Chen, Keng-Wei
  Hsu, and Yun-Nung Chen,
\newblock ``Slot-gated modeling for joint slot filling and intent prediction,''
\newblock in {\em Proc. of NAACL}, 2018.

\bibitem{li2018self}
Changliang Li, Liang Li, and Ji~Qi,
\newblock ``A self-attentive model with gate mechanism for spoken language
  understanding,''
\newblock in {\em Proc. of EMNLP}, 2018.

\bibitem{qin-etal-2019-stack}
Libo Qin, Wanxiang Che, Yangming Li, Haoyang Wen, and Ting Liu,
\newblock ``A stack-propagation framework with token-level intent detection for
  spoken language understanding,''
\newblock in {\em Proc. of EMNLP}, Nov. 2019.

\bibitem{NIPS2017_7181}
Ashish Vaswani, Noam Shazeer, Niki Parmar, Jakob Uszkoreit, Llion Jones,
  Aidan~N Gomez, $\L~ukasz$ Kaiser, and Illia Polosukhin,
\newblock ``Attention is all you need,''
\newblock in {\em NIPS}. 2017.

\bibitem{cui-zhang-2019-hierarchically}
Leyang Cui and Yue Zhang,
\newblock ``Hierarchically-refined label attention network for sequence
  labeling,''
\newblock in {\em Proc. of EMNLP}, 2019.

\bibitem{coucke2018snips}
Alice Coucke, Alaa Saade, Adrien Ball, Th{\'e}odore Bluche, Alexandre Caulier,
  David Leroy, Cl{\'e}ment Doumouro, Thibault Gisselbrecht, Francesco
  Caltagirone, Thibaut Lavril, et~al.,
\newblock ``Snips voice platform: an embedded spoken language understanding
  system for private-by-design voice interfaces,''
\newblock {\em arXiv preprint arXiv:1805.10190}, 2018.

\bibitem{devlin2018bert}
Jacob Devlin, Ming-Wei Chang, Kenton Lee, and Kristina Toutanova,
\newblock ``Bert: Pre-training of deep bidirectional transformers for language
  understanding,''
\newblock {\em arXiv preprint arXiv:1810.04805}, 2018.

\bibitem{hochreiter1997long}
Sepp Hochreiter and J{\"u}rgen Schmidhuber,
\newblock ``Long short-term memory,''
\newblock {\em Neural computation}, vol. 9, no. 8, 1997.

\bibitem{ba2016layer}
Jimmy~Lei Ba, Jamie~Ryan Kiros, and Geoffrey~E. Hinton,
\newblock ``Layer normalization,'' 2016.

\bibitem{kim-2014-convolutional}
Yoon Kim,
\newblock ``Convolutional neural networks for sentence classification,''
\newblock in {\em Proc. of EMNLP}, Oct. 2014.

\bibitem{e-etal-2019-novel}
Haihong E, Peiqing Niu, Zhongfu Chen, and Meina Song,
\newblock ``A novel bi-directional interrelated model for joint intent
  detection and slot filling,''
\newblock in {\em Proc. of ACL}, 2019.

\bibitem{liu-etal-2019-cm}
Yijin Liu, Fandong Meng, Jinchao Zhang, Jie Zhou, Yufeng Chen, and Jinan Xu,
\newblock ``{CM}-net: A novel collaborative memory network for spoken language
  understanding,''
\newblock in {\em Proc. of EMNLP}, 2019.

\bibitem{hemphill1990atis}
Charles~T Hemphill, John~J Godfrey, and George~R Doddington,
\newblock ``The atis spoken language systems pilot corpus,''
\newblock in {\em Speech and Natural Language: Proceedings of a Workshop Held
  at Hidden Valley, Pennsylvania, June 24-27, 1990}, 1990.

\bibitem{Pennington2014Glove}
Jeffrey Pennington, Richard Socher, and Christopher Manning,
\newblock ``Glove: Global vectors for word representation,''
\newblock in {\em Proc. of EMNLP}, 2014.

\bibitem{kingma-ba:2014:ICLR}
Diederik~P Kingma and Jimmy Ba,
\newblock ``Adam: A method for stochastic optimization,''
\newblock {\em arXiv preprint arXiv:1412.6980}, 2014.

\bibitem{haffner2003optimizing}
Patrick Haffner, Gokhan Tur, and Jerry~H Wright,
\newblock ``Optimizing svms for complex call classification,''
\newblock in {\em In Proc. of ICASSP}, 2003.

\bibitem{sarikaya2011deep}
Ruhi Sarikaya, Geoffrey~E Hinton, and Bhuvana Ramabhadran,
\newblock ``Deep belief nets for natural language call-routing,''
\newblock in {\em Proc. of ICASSP}, 2011.

\bibitem{raymond2007generative}
Christian Raymond and Giuseppe Riccardi,
\newblock ``Generative and discriminative algorithms for spoken language
  understanding,''
\newblock in {\em Eighth Annual Conference of the International Speech
  Communication Association}, 2007.

\bibitem{xu2013convolutional}
Puyang Xu and Ruhi Sarikaya,
\newblock ``Convolutional neural network based triangular crf for joint intent
  detection and slot filling,''
\newblock in {\em Proc. of ASRU}, 2013.

\bibitem{yao2014spoken}
Kaisheng Yao, Baolin Peng, Yu~Zhang, Dong Yu, Geoffrey Zweig, and Yangyang Shi,
\newblock ``Spoken language understanding using long short-term memory neural
  networks,''
\newblock in {\em SLT}, 2014.

\bibitem{zhao2019neural}
Sendong Zhao, Ting Liu, Sicheng Zhao, and Fei Wang,
\newblock ``A neural multi-task learning framework to jointly model medical
  named entity recognition and normalization,''
\newblock in {\em Proc. of AAAI}, 2019.

\bibitem{zhang-etal-2018-neural}
Rui Zhang, C{\'\i}cero Nogueira~dos Santos, Michihiro Yasunaga, Bing Xiang, and
  Dragomir Radev,
\newblock ``Neural coreference resolution with deep biaffine attention by joint
  mention detection and mention clustering,''
\newblock in {\em Proc. of ACL}, July 2018.

\bibitem{wang2018bi}
Yu~Wang, Yilin Shen, and Hongxia Jin,
\newblock ``A bi-model based rnn semantic frame parsing model for intent
  detection and slot filling,''
\newblock in {\em Proc. of NAACL}, 2018.

\end{thebibliography}

\end{document}